\pdfoutput=1

\documentclass[11pt]{article}

\usepackage{ACL}

\usepackage{times}
\usepackage{latexsym}

\usepackage[T1]{fontenc}

\usepackage[utf8]{inputenc}

\usepackage{microtype}

\usepackage{inconsolata}

\usepackage{lipsum}  
\usepackage{tabularx}
\usepackage{booktabs}

\usepackage{microtype}
\usepackage{type1cm}
\usepackage[american]{babel}
\usepackage{amssymb}
\usepackage{multirow}
\usepackage{xspace}
\DeclareMathAlphabet{\mathcalbf}{OMS}{pzc}{b}{n}
\usepackage{enumitem}
\usepackage{colortbl}
\usepackage{enumitem}

\RequirePackage{color}
\definecolor{darkgray}{gray}{0.40}
\definecolor{mediumgray}{gray}{0.60}
\definecolor{lightgray}{gray}{0.95}
\definecolor{ultralightgray}{gray}{0.98}
\definecolor{forestgreen}{rgb}{0.133, 0.545, 0.133}
\definecolor{orange}{rgb}{1, 0.86, 0.74}
\definecolor{lightergreen}{rgb}{0.95, 1, 0.88}

\usepackage{graphicx}
\DeclareGraphicsExtensions{.pdf,.ai,.jpg,.png}
\setkeys{Gin}{pagebox=artbox}
\graphicspath{{./bea24-essay-feedback-generation-figures/}}

\newcommand{\bsfigure}[3][]{%
    \begin{figure}[t]
        \centering
        \includegraphics[#1]{#2}
        \caption{#3}\label{#2}%
    \end{figure}
}

\newcommand{\hwfigure}[3][t!]{%
    \begin{figure*}[#1]
        \centering
        \includegraphics[scale=1.0]{#2}
        \caption{#3}\label{#2}%
    \end{figure*}
}

\RequirePackage{type1cm}
\RequirePackage{color}
\RequirePackage{soul}
\setstcolor{blue}
\definecolor{violet}{rgb}{0.5,0.0,0.5}

\newsavebox\bscombox
\newcommand{\bscom}[3][]{%
    \sbox{\bscombox}{\fontsize{8}{9}\selectfont#1#2#3}
    \noindent
    \st{#2}{\selectfont
        \color{blue}#3\ifx\\#1\\\else{\fontsize{8}{9}\selectfont\color{violet}[#1]}\fi
    }
}

\begin{document}


\title{Exploring LLM Prompting Strategies for \\ Joint Essay Scoring and Feedback Generation}

\author{
	Maja Stahl \\
	Leibniz University Hannover \\
	\texttt{m.stahl@ai.uni-hannover.de} \And
	Leon Biermann \\
	Leibniz University Hannover \\
	\texttt{leon.biermann@stud.uni-hannover.de} \AND
	Andreas Nehring \\
	Leibniz University Hannover \\
	\texttt{nehring@idn.uni-hannover.de} \And
	Henning Wachsmuth \\
	Leibniz University Hannover \\
	\texttt{h.wachsmuth@ai.uni-hannover.de}
}


\maketitle

\begin{abstract}
Individual feedback can help students improve their essay writing skills. However, the manual effort required to provide such feedback limits individualization in practice. Automatically-generated essay feedback may serve as an alternative to guide students at their own pace, convenience, and desired frequency. Large language models (LLMs) have demonstrated strong performance in generating coherent and contextually relevant text. Yet, their ability to provide helpful essay feedback is unclear. This work explores several prompting strategies for LLM-based zero-shot and few-shot generation of essay feedback. Inspired by Chain-of-Thought prompting, we study how and to what extent automated essay scoring (AES) can benefit the quality of generated feedback. We evaluate both the AES performance that LLMs can achieve with prompting only and the helpfulness of the generated essay feedback. Our results suggest that tackling AES and feedback generation jointly improves AES performance. However, while our manual evaluation emphasizes the quality of the generated essay feedback, the impact of essay scoring on the generated feedback remains low ultimately.
\end{abstract}

\newcommand{\bspubtag}{
	\vspace*{-18.65cm}\hspace*{-0.5cm}
	{\fontsize{6}{8}\selectfont%
		\renewcommand{\arraystretch}{0.9}
		\begin{tabular}{l}
			Accepted to the Workshop on Innovative Use of NLP for Building Educational Applications: BEA 2024
		\end{tabular}
}}

\bspubtag
\vspace*{17.5cm}\hspace*{0.5cm}

\section{Introduction}
\label{sec:introduction}

\bsfigure{essay-feedback-example}{Exemplary student essay on library censorship from the ASAP dataset \cite{hamner-etal-2012-hewlett} along with feedback and essay score generated by one of the methods evaluated in this paper. Explicit connections of the feedback to essay parts are color-coded.}

Essay writing is a valuable skill that empowers students to communicate effectively, to think critically, and to express their opinions clearly \cite{bangert-drowns-etal-2004-effects,liu-etal-2014-assessing,schendel-tolmie-2016-beyond}. However, learning to write essays of high quality can be challenging \cite{peloghitis-2017-difficulties,febriani-2022-writing}. Individual feedback to essays is thus critical, so students may recognize and overcome their weaknesses \cite{riddell-2015-performance}. However, manually writing essay feedback is time- and labor-intensive. Given educators' limited time, this often renders real individual feedback infeasible in practice. As an alternative, automated essay writing support could benefit students by offering guidance at their own pace and convenience.

Towards supporting essay writing automatically, much research has focused on quality assessment, mostly in the form of \emph{automated essay scoring} (AES; \citealp{ke-ng-2019-automated,ramesh-sanampudi-2022-automated}). Most existing AES approaches perform a holistic scoring, summarizing the quality of an essay with a single score \cite{alikaniotis-etal-2016-automatic,vajjala-2018-automated,uto-etal-2020-neural,wang-etal-2023-aggregating}. Alternatively, specific essay quality aspects may be assessed, such as coherence \cite{li-etal-2018-coherence,farag-etal-2018-neural}, grammar \cite{tambe-etal-2022-automated}, and organization \cite{persing-etal-2010-organization,rahimi-etal-2015-incorporating}. While AES helps assess essay quality and monitor writing skill progress, most approaches cannot explain why a score was predicted, nor guide the student in improving the essay. 

Another prominent line of research towards writing support focuses on identifying and correcting grammatical errors \cite{imamura-etal-2012-grammar,bryant-etal-2017-automatic,rozovskaya-roth-2019-grammar,grundkiewicz-etal-2019-neural}. However, studies in educational research show that computer-based learning systems lead to higher learning outcomes if elaborated feedback is provided that provides explanations instead of only pointing to errors or providing the solution \cite{van-der-kleij-etal-2015-effects}. 
Therefore, \citet{nagata-2019-toward} introduced the task of \emph{feedback comment generation} in NLP: Given a learner text with a grammatical error, automatically generate a comment with hints and explanations to guide their correction process. \citet{song-etal-2023-gee} extended this task by generating explanations for a broader range of grammatical error types using large language models (LLMs). However, these tasks operate only on the sentence level and are limited to grammatical errors. Generating feedback on the essay level by addressing not only grammatical errors but the essay as a whole remains relatively unexplored. 

To foster research in this direction, we tackle the task of \emph{essay feedback generation}: Given a student essay, automatically generate textual feedback that helps students improve their essays. An example is shown in Figure~\ref{essay-feedback-example}. Building on the strong abilities of LLMs in many text-generation tasks, this work examines how well LLMs can generate essay feedback by exploring various prompting strategies in zero- and few-shot settings. Inspired by Chain-of-Thought prompting \cite{wei-etal-2022-chain}, we study whether AES can benefit the performance of essay feedback generation and vice versa. 

Our experiments suggest that generating essay feedback by explaining the predicted essay score improves the scoring performance on the widely-used ASAP dataset \cite{hamner-etal-2012-hewlett}. For essay feedback generation, we deem helpfulness to be the most important quality criterion. Helpful essay feedback should point out and explain mistakes made in an essay in a precise and easy way for students to understand \cite{shute-2008-focus,hattie-timperley-2007-power}. We evaluate the helpfulness automatically and manually. Due to the lack of ground-truth essay feedback, we propose using LLMs to automatically judge the essay feedback's helpfulness, which turns out to correlate well with human helpfulness judgments.
Our manual evaluation also reveals that the generated essay feedback is deemed helpful for students to improve their essay writing skills. However, the impact of scoring the essay remains low ultimately.
Altogether, this paper's main contributions are:
\begin{itemize}[itemsep=0pt]
    \item A comparison of several LLM prompting strategies for automated essay scoring
    \item An approach and task-specific automatic evaluation strategy for essay feedback generation using LLM prompting
    \item Empirical insights into the influence of automated essay scoring on generating essay feedback and vice versa\footnote{The code used for our experiments can be found under \url{https://github.com/webis-de/BEA-24}.}
\end{itemize}

\section{Related Work}
\label{sec:related-work}

Essay writing is a central task in education to evaluate various skills of students, including logical thinking, critical reasoning, and creativity \cite{liu-etal-2014-assessing,schendel-tolmie-2016-beyond}. However, manual essay grading is time-consuming and not always consistent within and across raters \cite{kassim-2011-judging,eckes-2015-introduction}. Automated essay scoring (AES) aims to alleviate these issues, reducing the effort of graders and, ideally, making grading more consistent and reliable \cite{ke-ng-2019-automated,uto-2021-review,ramesh-sanampudi-2022-automated}.

While extensive research exists on AES \cite{ke-ng-2019-automated,ramesh-sanampudi-2022-automated}, assessing all important quality aspects (known as \emph{traits}), including the relevance of an essay's content to the prompt, the development of ideas, cohesion, coherence, and more remains challenging \cite{ramesh-sanampudi-2022-automated}. Only few works focus on scoring multiple traits at once \cite{mathias-bhattacharyya-2020-neural,hussein-2020-trait}. Instead, the majority of AES research targets holistic essay scoring, that is, summarizing the essay quality in a single score \cite{alikaniotis-etal-2016-automatic,cozma-etal-2018-automated,vajjala-2018-automated,wang-etal-2023-aggregating}.

State-of-the-art approaches to AES can be divided by their use of the available data into full-data and few- or zero-shot settings \cite{tao-etal-2022-aesprompt}. In the full-data setting, where all labeled data is used for training, most approaches fine-tune pretrained language models, such as BERT \cite{devlin-etal-2019-bert}. \citet{yang-etal-2020-enhancing} proposed solving the task by combining essay scoring and essay ranking, fine-tuning BERT using multiple losses simultaneously. Extending this idea, \citet{xie-etal-2022-automated} combined regression and ranking into a single loss. Rather than fine-tuning a language model, \citet{tao-etal-2022-aesprompt} designed two self-supervised constraints for learning a multi-layer embedding, which prepends the input to a frozen BERT model. They evaluate their approach in the full-data and one-shot setting, outperforming a fine-tuned BERT in the latter. To explore the potential of large language models (LLMs), \citet{mizumoto-eguchi-2023-exploring} prompted GPT-3.5 to score the student essays from the TOEFL11 dataset \cite{blanchard-etal-2013-toefl11} in a zero-shot setting, indicating promising scoring performance.

The most straightforward way to provide more detailed feedback for an essay than a holistic score is trait scoring \cite{jong-etal-2023-review}, that is, to evaluate an essay for different quality aspects. However, the reasoning behind an assigned trait score usually remains unknown to the student. Therefore, \citet{kumar-boulanger-2020-explainable} adopted explainability methods to explain how input features to an AES system influence the trait scores for an essay. While this provides more insights, the pedagogical quality and impact on writing performance remain questionable if no feedback is given together with the scores \cite{kumar-boulanger-2020-explainable}.

Specific feedback generation tasks have been addressed in educational NLP. \citet{nagata-2019-toward} introduced {\it feedback comment generation} to explain grammatical errors to a learner on the sentence level. This task has been tackled by combining retrieval and text generation \cite{hanawa-etal-2021-exploring,ihori-etal-2023-retrieval}, by identifying different feedback types \cite{stahl-wachsmuth-2023-identifying}, by augmenting the dataset \cite{babakov-etal-2023-error,behzad-etal-2023-sentence}, and by correcting the error \cite{jimichi-etal-2023-feedback,koyama-etal-2023-tokyo}, all using fine-tuned language models. For a wider range of grammatical error types, \citet{song-etal-2023-gee} used the LLM GPT-4 to first identify the necessary corrective edit before generating a grammar error explanation using one-shot prompting. In the educational domain, \citet{meyer-etal-2024-using} showed that LLM-based writing feedback, generated using a single handcrafted prompt, positively impacts students' text revisions, motivation, and positive emotions.

So far, however, the generation of textual feedback on complete student essays has, to our knowledge, received very little attention. All generation approaches mentioned above operate on the sentence level and explain grammatical errors only, while our work aims to address all aspects of student essays that may need improvement. The only other work on essay feedback generation tackled the task using Chain-of-Thought prompting using zero-shot learning \cite{han-etal-2023-fabric}. The resulting feedback was deemed to be more helpful than the feedback generated using standard prompting, as evaluated by humans.

Motivated by these promising results and the positive effects of LLM-based writing feedback on students, we go beyond previous work by comparing the effectiveness of different prompting strategies for essay feedback generation. We also study how and to what extent AES can benefit essay feedback generation (and vice versa) by addressing the tasks jointly. Following the educational literature on feedback, we aim to generate essay feedback that is specific and elaborate \cite{shute-2008-focus} while assessing the current state and instructing on how to improve to achieve the goals \cite{hattie-timperley-2007-power}.

\section{Approach}
\label{sec:approach}

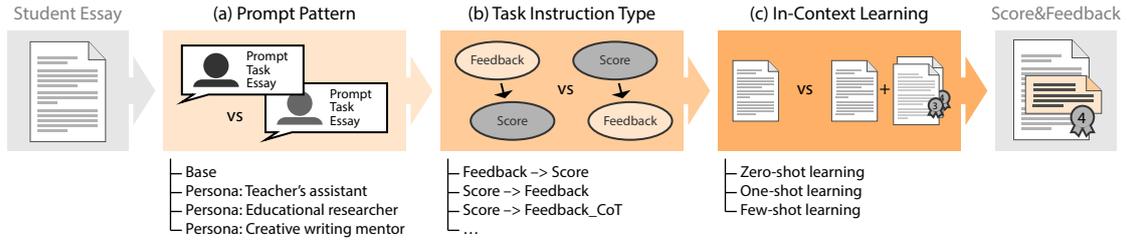
\hwfigure{prompting-approach}{Overview of the main points of variation in our approach to predict a score and to generate feedback for a student essay: (a) Prompt pattern: Use of the base pattern or persona-specific pattern; (b) Task instruction type: Tasks to be tackled and their ordering; (c) In-context learning approach: Number of examples to learn from.}

This section describes our approach to essay feedback generation. We propose to tackle essay scoring and feedback generation jointly in order to study how and to what extent AES can benefit essay feedback generation and vice versa. By comparing different prompting strategies for large language models (LLMs), we explore how well the tasks can be solved using in-context learning. 

In particular, we test different prompting strategies by systematically varying three main aspects of the prompts, as visualized in Figure~\ref{prompting-approach}: (a) the \emph{prompt pattern}, which defines the context and layout; (b) the \emph{task instruction type}, which sets the ordering and phrasing of the tasks to be tackled; and (c) the \emph{in-context learning} approach, which specifies the number of examples provided.

\begin{table}[t]
    \centering
    \small
    \setlength{\tabcolsep}{2pt}
    \begin{tabular}{p{0.983\columnwidth}}
        \toprule
        {\bf Base:}
        You are given an essay written by a student and the corresponding prompt for the 7th to 10th grade student. \\
        \#\#\#\# Prompt: ``\{{\it essay\_prompt}\}'' \\
        \#\#\# Task: \{{\it task\_instruction}\}\\
        \#\#\#\# Student essay: ``\{{\it essay}\}'' \\
        \midrule        
        {\bf Teacher's Assistant}:
        Imagine you are a teacher's assistant in a middle school tasked with reviewing a 7th to 10th grade student's essay. You have the essay and the prompt that was given to the student. \\
        \#\#\#\#  Original Prompt Provided to Student: ``\{{\it essay\_prom.}\}'' \\
        \#\#\#  Review Task: \{{\it task\_instruction}\}\\
        \#\#\#\# Student's Essay for Review: ``\{{\it essay}\}'' \\
        \midrule        
        {\bf Educational Researcher:}
        You are part of an educational research team analyzing the writing skills of students in grades 7 to 10. You have been given a student's essay and the prompt they responded to. \\
        \#\#\#\# Essay Prompt: ``\{{\it essay\_prompt}\}'' \\
        \#\#\# Analysis Task: \{{\it task\_instruction}\}\\
        \#\#\#\# Analyzed Student Essay: ``\{{\it essay}\}'' \\
        \midrule
        {\bf Creative Writing Mentor:}
        You are a creative writing mentor evaluating a piece written by a student in grades 7 to 10. The student's work is based on a specific prompt. \\
        \#\#\#\#  Creative Prompt Given: ``\{{\it essay\_prompt}\}'' \\
        \#\#\#  Critique Instructions: \{{\it task\_instruction}\}\\
        \#\#\#\#  Student's Creative Piece: ``\{{\it essay}\}'' \\
        \bottomrule
    \end{tabular} 
    \caption{Prompt patterns: Base pattern and all persona patterns. Brackets indicate placeholders that are filled respectively during the experiments. We removed model-specific pre-/suffixes and line breaks for illustration.}
    \label{tab:patterns}
\end{table}

\subsection{Prompt Patterns}

We compare two different kinds of prompt patterns, which define the context and format of the prompt: (i) a \emph{base pattern} and (ii) different \emph{persona patterns}. All prompt patterns are displayed in Table~\ref{tab:patterns}. 

\paragraph{Base Pattern}
The base pattern simply gives the general context and defines the layout and order in which the \emph{essay prompt} (i.e., the task given to the learner writing the essay), the \emph{task instruction}, as defined by the used task instruction type, and the current learner \emph{essay} will be presented to the model. All inputs are indicated by markdown headings. 

\paragraph{Persona Patterns}
These prompt patterns are inspired by persona prompting \cite{white-etal-2023-prompt}, giving the LLM a persona or role to play when generating output. This aims to implicitly define the expected type of output. For our task, we compare the three personas, namely, \emph{teacher's assistant}, \emph{educational researcher}, and \emph{creative writing mentor}, by altering the context given in the prompt pattern.

\subsection{Task Instruction Types}

The task instruction type defines the tasks to be tackled along with their ordering. We differentiate between tackling (i)~only essay \emph{scoring}, (ii)~essay \emph{scoring and feedback} generation, and (iii)~only essay \emph{feedback} generation. This way, we can measure the influence that essay scoring has on feedback generation, and vice versa.
We explore the following task instruction types for our tasks:
\begin{itemize}[itemsep=0pt]
    \item 
    {\it Scoring}. Instruct to score the student essay on a given score range. This serves as a baseline for assessing the essay scoring performance.
    \item 
    {\it Feedback}. Instruct to generate essay feedback for the student writer. This serves as a baseline for assessing the feedback performance.
    \item 
    {\it Scoring$\rightarrow$Feedback}. Instruct to score the essay and then generate feedback for the student writer. This measures the influence of essay scoring on the feedback performance. 
    \item 
    {\it Feedback$\rightarrow$Scoring}. Instructs to first generate feedback before scoring the essay. This evaluates whether feedback generation helps to predict the correct essay score.
    \item 
    {\it Scoring$\rightarrow$Feedback\_CoT}. Instruct to score the essay and to then generate feedback using zero-shot Chain-of-Thought (CoT) prompting, that is, to add the phrase ``Let's think step by step.'', which has been shown to increase LLM's reasoning performance \cite{kojima-etal-2022-large}. This might benefit the reasoning needed in feedback generation.
    \item 
    {\it Feedback\_dCoT$\rightarrow$Scoring}. Instruct to first analyze the essay quality using the rubric, to then generate feedback, and to finally score the essay. This is a more detailed variation of CoT that provides task-specific steps to follow before arriving at the final essay score.
    \item 
    {\it Scoring$\rightarrow$Explanation}. Instruct to score the essay and to then generate an explanation for the predicted score. This explores whether score explanations as a form of feedback relate to asking for essay feedback specifically.
    \item 
    {\it Explanation$\rightarrow$Scoring}. Instruct to analyze the essay, to then first generate an explanation for an essay score that, in turn, should be generated at the end. This avoids that the LLM predicts an incorrect score and then generates an explanation justifying the incorrect score, as observed by \citet{ye-durrett-2022-unreliability}.
\end{itemize}

\begin{table}[t]
    \centering
    \small
	\renewcommand{\arraystretch}{0.95}
    \setlength{\tabcolsep}{2pt}
    \begin{tabular}{lp{0.86\columnwidth}}
        \toprule
        {\bf Score}	& \bf Description \\
        \midrule
        {3} 			& The response demonstrates an understanding of the complexities of the text.\\
       				& -- Addresses the demands of the question\\
        				& -- Uses expressed and implied information from the text\\
       				& -- Clarifies and extends understanding beyond the literal\\ 
        \addlinespace
        {2} 			& The response demonstrates a partial or literal understanding of the text.\\
        				& -- Addresses the demands of the question, although may not develop all parts equally\\
       				& -- Uses some expressed or implied information from the text to demonstrate understanding\\
       				& -- May not fully connect the support to a conclusion or assertion made about the text(s)\\
        \addlinespace
        {1}		 	& The response shows evidence of a minimal understanding of the text.\\
        				& -- May show evidence that some meaning has been derived from the text\\
        				& -- May indicate a misreading of the text or the question\\
        				& -- May lack information or explanation to support an understanding of the text in relation to the question\\
        \addlinespace
        {0}		 	& The response is completely irrelevant or incorrect, or there is no response.\\
        \bottomrule
    \end{tabular} 
    \caption{Exemplary rubric from essay set 3 of the ASAP dataset \cite{hamner-etal-2012-hewlett}. The rubrics are provided as additional information within  the task instructions.}
    \label{tab:rubric-example}
\end{table}

\begin{table}[t]
    \centering
    \small
    \setlength{\tabcolsep}{2pt}
    \begin{tabular}{p{0.983\columnwidth}}
        \toprule
        {\bf Scoring:}
        Given this essay that was written for the given prompt, grade the essay using those ranges: \{{\it scoring\_range}\}. \\ 
        \midrule
        {\bf \bf Feedback:}
        Analyze the given essay using the following rubric: \{{\it rubric}\}. Provide comprehensive feedback for the student that helps them to achieve better grades in the future. \\      
        \midrule
        {\bf Scoring$\rightarrow$Feedback:}
        Grade the given essay using the following rubric: \{{\it rubric}\}. Use those score ranges: \{{\it scoring\_range}\}. Provide comprehensive feedback for the student that helps them to achieve better grades in the future. \\  
        \midrule
        {\bf Feedback\_dCoT$\rightarrow$Scoring:}
        Analyze the given essay using the following rubric and give helpful feedback to the student: \{{\it rubric}\}. Use those score ranges: \{{\it scoring\_range}\}. Let's think step by step. First, analyze the quality of the essay in terms of the given rubric. Then, give feedback to the student that explains their mistakes and errors and additionally gives them tips to avoid them in the future. As a final step, output the score at the end. \\ 
        \midrule
        {\bf Scoring$\rightarrow$Explanation:}
        Grade the given essay using the following rubric: \{{\it rubric}\}. Use those score ranges: \{{\it scoring\_range}\}. Provide an explanation for your score as well. \\
        \bottomrule
    \end{tabular} 
    \caption{Task instruction types: Examples of the initial, manually written task instructions for five types. Brackets indicate placeholders that are filled with the respective information during the experiments.}
    \label{tab:task-instruction-types}
\end{table}
\begin{table}[t]
    \centering
    \small
    \setlength{\tabcolsep}{2pt}
    \begin{tabular}{p{0.983\columnwidth}}
        \toprule
        {\bf One-shot Example: } Essay: ``\{{\it essay}\}'' \\
        \addlinespace
        Reasoning: This is a minimally-developed response with inadequate support and detail. The writer takes the position that computers can be harmful to the eyes and then addresses eye damage to three groups of people (kids, teens, adults). A few specific details are included (sensitive eyes, MySpace), but elaboration is minimal. Some organization is demonstrated but few transitions are used. Overall, the response is sufficiently developed to move into the score point `3' range. \\
        \addlinespace
        Scores: \{Overall: 3\}\\
        \bottomrule
    \end{tabular} 
    \caption{One-shot example consisting of a student essay, a manually written score justification, and the assigned score. The data is taken from the scoring guidelines for essay set 1 of the ASAP dataset \cite{hamner-etal-2012-hewlett}.}
    \label{tab:one-shot-example}
\end{table}

Task instructions for essay scoring provide the {\it scoring range} that should be used, while those for feedback generation provide the {\it rubric}, that is, guidelines including a short description for essays of each quality level and typical elements of such. An exemplary {\it rubric} can be seen in Table~\ref{tab:rubric-example}.

Since the performance of LLMs is sensitive to the exact wording of a prompt \cite{leidinger-etal-2023-language}, we create a total of four {\it task instructions} for each task instruction type by instructing ChatGPT \cite{openai-2023-chatgpt} to generate three paraphrases of each initial, manually written task instruction. Examples of the latter can be seen in Table~\ref{tab:task-instruction-types}. We provide all task instructions in Appendix \ref{app:task-instructions}.

\subsection{In-Context Learning}

As final point of variation of our approach, we explore how providing one or multiple exemplary essays, together with their score and a reasoning for the score, helps with essay scoring and feedback generation. The data comes from additional material given to human raters. We argue that the reasoning of the score may help with essay scoring, but could also be seen as a form of feedback and may benefit that task as well. We compare (i) \emph{zero-shot}, (ii) \emph{one-shot}, and (iii) \emph{few-shot} learning. 

For one-shot, we randomly select an essay with a medium score, as the one in Table~\ref{tab:one-shot-example}. For few-shot, we first randomly select examples among the essays with the best and worst scores before covering the other scores. Due to the limited context length, we restrict the prompt to 5,120 characters and select as many examples that fit this limitation as possible.%
\footnote{For the few-shot variation, the described example selection process led to 3, 2, 4, 5, 8, 6, 4 and 2 examples for the essay sets 1 to 8 respectively. The differences are due to the variation in essay and reasoning length per essay set.}

\section{Data}
\label{sec:data}

\begin{table}[t]
    \centering
    \small
    \setlength{\tabcolsep}{1.9pt}
    \begin{tabular}{lrrrrrrrrrr}
        \toprule
        \bf Pattern && \multicolumn{9}{c}{\bf Essay Set} \\
        \cmidrule{3-11}
        && \multicolumn{1}{c}{\bf 1} & \multicolumn{1}{c}{\bf 2} & \multicolumn{1}{c}{\bf 3} & \multicolumn{1}{c}{\bf 4} & \multicolumn{1}{c}{\bf 5} & \multicolumn{1}{c}{\bf 6} & \multicolumn{1}{c}{\bf 7} & \multicolumn{1}{c}{\bf 8} & \bf Mean \\
        \midrule
        Base && .495 & .532 & .405 & .495 & .497 & .601 & .436 & .377 & .480 \\
        TA && \bf .536 & \bf .603 & .408 & .499 & .512 & \bf .625 & .443 & .439 & .508 \\
        ER && .436 & .554 & \bf .460 & \bf .560 & \bf .553 & .620 & .418 & \bf .467 & \bf .509 \\
        CWM && .484 & .588 & .382 & .434 & .507 & .596 & \bf .471 & .352 & .477 \\
        \bottomrule
    \end{tabular}
    \caption{Essay scoring results: Average QWK over all task instructions using zero-shot learning for each prompt pattern: base, teacher's assistant (TA), educational researcher (ER), and creative writing mentor (CWM). We report the performance for each of the eight essay sets as well as the mean QWK over all sets.}
    \label{tab:qwk-results-patterns}
\end{table}
\begin{table*}[t]
	\centering
	\small
	\setlength{\tabcolsep}{5pt}
	\begin{tabular}{lcccccccccccr}
		\toprule
 	\bf Task Instruction Type && \multicolumn{9}{c}{\bf Essay Set} && \bf Unscored \\
        \cmidrule{3-11}
        && \bf 1 & \bf 2 & \bf 3 & \bf 4 & \bf 5 & \bf 6 & \bf 7 & \bf 8 & \bf Mean \\
		\midrule     
        Scoring && .448 & .585 & .479 & .596 & .557 & .649 & .438 & .481 & .529 && 1 \\
        Scoring$\rightarrow$Feedback && .510 & \bf .615 & .439 & .530 & .489 & .621 & .449 & .481 & .517 && 1 \\
        Feedback$\rightarrow$Scoring && .388 & .561 & .484 & .600 & \bf .622 & .630 & .385 & \bf .545 & .527 && 16 \\
        Scoring$\rightarrow$Feedback\_CoT && .538 & .595 & .422 & .494 & .530 & .635 & .458 & .477 & .519 && 19 \\
        Feedback\_dCoT$\rightarrow$Scoring && \bf .546 & .564 & .424 & .558 & .581 & .628 & .477 & .489 & \bf .533 && 37 \\
        Scoring$\rightarrow$Explanation && .466 & .580 & .472 & .565 & .541 & .639 & .420 & .417 & .513 && 0 \\
        Explanation$\rightarrow$Scoring && .470 & .553 & \bf .488 & .636 & .571 & \bf .675 & .384 & .484 & \bf .533 && 2 \\  
		\bottomrule
	\end{tabular}
	\caption{Essay scoring results:  QWK for the best approach variation per task instruction type in the zero-shot setting. We report the performance per essay set and the average over essay sets. The best results per column are bold.}
	\label{tab:qwk-results}
\end{table*}

Multiple AES datasets are available, with the Automated Student Assessment Prize’s (ASAP) dataset \cite{hamner-etal-2012-hewlett} being the most widely used. It comprises 12,980 essays written by school students in grades 7 to 10. All essays were scored manually by two raters. The essays are divided into eight essay sets. The essay sets differ by the essay prompt, i.e., the task description they were written for, the scoring range, and the rubric used by the raters as annotation guidelines. The rubrics provide a short description for essays of each quality level and typical elements of such essays. 

Since for the introduced task of essay feedback generation, no parallel dataset is available yet, we use the ASAP dataset as input data and evaluate the generated feedback without supervision.

\section{Evaluation}
\label{sec:evaluation}

We evaluate the performance of a large language model (LLM) by comparing the proposed prompting strategies on the two tasks: essay scoring and feedback generation. First, we assess the scoring performance and, then, we both automatically and manually evaluate the generated feedback in terms of the helpfulness for the student writer. We aim to study the effects of tackling essay scoring and feedback generation jointly, as well as explore how well LLMs can solve both tasks using prompting.

\subsection{Essay Scoring}

We compare the proposed prompt patterns, task instruction types, and in-context learning approaches, to evaluate the performance of an LLM on the essay scoring task. Also, we measure the influence of feedback generation on the scoring performance.

\paragraph{Approach}

We use the instruction-following recent LLM Mistral with 7B parameters ({\it Mistral-7B-Instruct-v0.2}, \citealp{jiang-etal-2023-mistral}) in our experiments, generating each output with greedy decoding.%
\footnote{Initial experiments on essay scoring with Llama-2 ({\it 7b-chat-hf} and {\it 13b-chat-hf}, \citealp{touvron-etal-2023-llama}) led to lower performance, which halted further testing with Llama-2.}
We found that instructing the model to generate the essay score in JSON format helps to extract the score from the generated text automatically.%
\footnote{If the score was not generated as instructed, we re-prompted the model to extract the score from its prior response. This was effective when a score was in the initial answer.}
Below, we report the number of essays that still did not receive a score ({\it Unscored}) and omit them from the performance calculation. 

\paragraph{Baselines}
As a baseline, we report the performance AES-Prompt \cite{tao-etal-2022-aesprompt}, which is, to our knowledge, the best-performing AES approach that is not fully fine-tuned on the ASAP dataset. As an upper bound, we also report the performance of R$^2$BERT \cite{yang-etal-2020-enhancing}, the state-of-the-art approach fully fine-tuned on the same dataset.

\paragraph{Experimental Setup}
We automatically assess the essay scoring performance using quadratic weighted kappa (QWK), the most widely adopted metric for automatic essay scoring \cite{ke-ng-2019-automated}. Since the test set of the ASAP dataset is not publicly available, we follow \citet{taghipour-ng-2016-neural} and apply their 5-fold cross-validation split. Since we perform no training, we only use the validation splits to create reasonable initial prompts and report the performance on the test splits.

\paragraph{Results}

Table~\ref{tab:qwk-results-patterns} presents the scoring performance for each prompt pattern. We report the average QWK of all task instructions using zero-shot learning to measure the influence of the prompt pattern on the scoring performance. Using the personas ``educational researcher'' (ER) and ``teacher's assistant'' (TA) seems beneficial for essay scoring, either of which performs best on all but one essay set, and ER best on average (mean QWK of .509). 

To evaluate the influence of the task instruction type, Table~\ref{tab:qwk-results} shows the performance of the best-performing approach variations per task instruction type. We report the combination of prompt pattern and task instruction that performed best on the validation set using zero-shot learning. The results suggest that instructing the LLM to first follow task-specific steps to analyze and give feedback ({\it Feedback\_dCoT$\rightarrow$Scoring}) as well as to first generate an explanation for the essay score ({\it Explanation$\rightarrow$Scoring}) particularly help with essay scoring. These two achieve the highest mean QWK (.553). In general, the variations that generate some form of feedback first perform better than their counterparts that perform scoring first.

Finally, we study the influence of in-context learning on the instruction type {\it Scoring$\rightarrow$Feedback} using the prompt pattern and task instruction that performs best on the validation split for a fair comparison to the baselines (Table~\ref{tab:qwk-results-shots}). The results indicate that giving examples of scored essays aid essay scoring. One-shot learning outperforms few-shot learning, but the effect is rather small. Our prompting approaches perform rather competitively to the strong baseline AES-Prompt \cite{tao-etal-2022-aesprompt}.

\subsection{Essay Feedback Generation}

As with essay scoring, we evaluate the generated feedback by comparing the prompt patterns, task instruction types, and in-context learning approaches. Our goal is to explore how well LLMs perform at generating helpful essay feedback and whether essay scoring can benefit the feedback generation.

\begin{table}[t]
    \centering
    \small
    \setlength{\tabcolsep}{1.8pt}
    \begin{tabular}{lccccccccc}
        \toprule
        \bf Context & \multicolumn{9}{c}{\bf Essay Set} \\
        \cmidrule{2-10}
        & \bf 1 & \bf 2 & \bf 3 & \bf 4 & \bf 5 & \bf 6 & \bf 7 & \bf 8 & \bf Mean \\
        \midrule
        Zero-shot & .510 & .615 & .439 & .530 & .489 & .621 & .449 & .481 & .517 \\
        One-shot & .565 & \bf .619 & .523 & .600 & .606 & .665 & .509 & .233 & .540 \\
        Few-shot & .558 & .586 & .515 & .586 & .618 & \bf .671 & .472 & .297 & .538 \\
        AES-Pro. & \bf .682 & .544 & \bf .590 & \bf .672 & \bf .701 & .622 & \bf .683 & \bf .620 & \bf .639 \\
        \midrule
        R$^2$BERT & .817 & .719 & .698 & .845 & .841 & .847 & .839 & .744 & .794 \\
        \bottomrule
    \end{tabular}
    \caption{Essay scoring results: QWK per in-context learning approach for {\it Scoring$\rightarrow$Feedback} using the best-performing prompt pattern and task instruction. The baseline {\it AES-Prompt} \cite{tao-etal-2022-aesprompt} also has one shot. {\it R$^2$BERT} \cite{yang-etal-2020-enhancing} is fully fine-tuned. }
    \label{tab:qwk-results-shots}
\end{table}

\paragraph{Approach} We continue using the large language model Mistral ({\it Mistral-7B-Instruct-v0.2}, \citealp{jiang-etal-2023-mistral}) for the essay feedback generation task since it performed well at the essay scoring task.

\paragraph{Automatic Evaluation}
Using LLMs to assess the quality of generated texts has been shown to be consistent with human expert annotations for some free-text generation tasks \cite{chiang-lee-2023-large}. Since there are no existing automatic metrics for assessing the quality of generated essay feedback, we follow previous work and use Mistral itself as well as Llama-2 ({\it Llama-2-13b-chat-hf}, \citealp{touvron-etal-2023-llama}) for the automatic part of our feedback evaluation. We instruct them to assign an overall helpfulness scores between 1 (not helpful) and 10 (very helpful) for each generated essay feedback. The used prompt can be found in Appendix~\ref{app:automatic-helpfulness}.%
\footnote{We also experimented with relative comparisons of feedback for automatic helpfulness assessment. However, the correlation to our manual helpfulness annotations was low.} 

Our evaluation focuses on helpfulness, which we deem to be the most important quality dimension for essay feedback. We anticipate that other quality aspects, such as faithfulness, are implicitly covered since irrelevant or incorrect feedback would not be helpful for the student author.

\paragraph{Automatic Results}

\begin{table}
    \centering
    \small
    \setlength{\tabcolsep}{3pt}
    \begin{tabular}{lcccc}
        \toprule
        \bf Prompt Pattern && \bf Mistral & \bf Llama-2 \\
        \midrule
        Base && 7.78 \scriptsize $\pm 0.53$ & 6.88 \scriptsize $\pm 0.18$ \\
        Teacher’s assistant (TA) && 7.90 \scriptsize $\pm 0.39$ & 6.84 \scriptsize $\pm 0.19$ \\
        Educational researcher (ER) && \bf 8.26 \scriptsize $\pm 0.23$ & 6.87 \scriptsize $\pm 0.18$ \\
        Creative writing mentor (CWM) && 7.83 \scriptsize $\pm 0.47$ & \bf 7.48 \scriptsize $\pm 0.85$ \\
        \bottomrule
    \end{tabular}
    \caption{Automatic feedback generation results: Average helpfulness scores predicted by Mistral and Llama-2 for each prompt pattern over all task instructions using zero-shot learning. The best result per column is bold.}
    \label{tab:automatic-absolute-helpfulness-patterns}
\end{table}

Table~\ref{tab:automatic-absolute-helpfulness-patterns} presents the assigned helpfulness scores for each prompt pattern, averaged over task instructions using zero-shot learning. Both LLMs deemed the feedback generated by a persona pattern to be most helpful, on average: the top helpfulness score is achieved by ER for Mistral (8.26) and CWM for Llama-2 (7.48).

\begin{table}[t]
    \centering
    \small
    \setlength{\tabcolsep}{6pt}
    \begin{tabular}{lccc}
        \toprule
        \bf Task Instruction Type && \bf Mistral & \bf Llama-2 \\ 
        \midrule
        Feedback && \bf 8.96 \scriptsize $\pm .25$ & \bf 7.31 \scriptsize $\pm .19$ \\
        Scoring$\rightarrow$Feedback && 8.04 \scriptsize $\pm .44$ & 7.15 \scriptsize $\pm .45$ \\
        Feedback$\rightarrow$Scoring && 8.27 \scriptsize $\pm .38$ & 7.27 \scriptsize $\pm .50$ \\
        Scoring$\rightarrow$Feedback\_CoT && 7.30 \scriptsize $\pm .63$ & 6.72 \scriptsize $\pm .41$ \\
        Feedback\_dCoT$\rightarrow$Scoring && 8.53 \scriptsize $\pm .66$ & 7.28 \scriptsize $\pm .55$ \\
        Scoring$\rightarrow$Explanation && 7.22 \scriptsize $\pm .45$ &  6.68 \scriptsize $\pm .40$ \\
        Explanation$\rightarrow$Scoring && 7.27 \scriptsize $\pm .63$ & 6.75 \scriptsize $\pm .36$ \\
        \bottomrule
    \end{tabular}
    \caption{Automatic feedback generation results: Average helpfulness scores predicted by Mistral or Llama-2 for each task instruction type over all task instructions and prompt patterns using zero-shot learning.}
    \label{tab:automatic-absolute-helpfulness-results}
\end{table}

To evaluate the influence of the task instruction type, Table~\ref{tab:automatic-absolute-helpfulness-results} shows the results per type, averaged over prompt patterns and task instructions using zero-shot learning. Both evaluation models gave the highest average scores to performing feedback generation only ({\it Feedback}). For the other task instruction types, the variations that generate some form of feedback first seem more helpful than their counterparts that perform scoring first.

\begin{table}[t]
    \centering
    \small
    \setlength{\tabcolsep}{5pt}
    \begin{tabular}{lcccc}
        \toprule
        \bf In-Context Learning && \bf Mistral & \bf Llama-2 \\
        \midrule
        Zero-shot learning && 8.04 \scriptsize $\pm .44$ & 7.15 \scriptsize $\pm .45$ \\
        One-shot learning && 8.39 \scriptsize $\pm .54$ & 7.28 \scriptsize $\pm .47$  \\
        Few-shot learning && \bf 8.42 \scriptsize $\pm .56$ & \bf 7.30 \scriptsize $\pm .46$ \\
        \bottomrule
    \end{tabular}
    \caption{Automatic feedback generation results: Average helpfulness scores predicted by Mistral or Llama-2 per in-context learning approach for {\it Scoring$\rightarrow$Feedback} over all prompt patterns and task instructions.}
    \label{tab:automatic-absolute-helpfulness-shots}
\end{table}

Finally, we study the influence of each in-context learning approach on the task instruction type {\it Scoring$\rightarrow$Feedback} on average over the prompt patterns and task instructions (Table~\ref{tab:automatic-absolute-helpfulness-shots}). The results suggest that the reasoning presented in the provided in-context examples positively impacts the feedback helpfulness. Although the effect is small, more examples help more.

\begin{table}[t]
    \centering
    \small
    \setlength{\tabcolsep}{3pt}
    \begin{tabular}{lcccccc}
        \toprule
        \bf Task Instruction Type && \multicolumn{1}{c}{\bf S1} & \multicolumn{1}{c}{\bf S2} & \multicolumn{1}{c}{\bf S3} & \multicolumn{1}{c}{\bf S4} & \multicolumn{1}{c}{\bf S5} \\
        \midrule
        Feedback && \bf 5.88 & \bf 5.71 & \bf 6.04 & \bf 5.75 & \bf 6.08 \\
        Feedback$\rightarrow$Scoring && 5.17 & 5.04 & 5.46 & 5.21 & 5.08 \\
        Feedback\_dCoT$\rightarrow$Scoring && 5.50 & 4.92 & 5.29 & 4.83 & 5.00 \\
        \bottomrule
    \end{tabular}
    \caption{Manual feedback generation results: Average scores assigned by the annotators for each approach for statements S1--S5 on a 7-point Likert scale (7 is best).}
    \label{tab:manual-results}
\end{table}

\paragraph{Manual Evaluation}
The proposed automatic evaluation only approximates the quality of the generated essay feedback. Therefore, we conducted a manual annotation study during which 12 annotators manually judged the feedback quality. All annotators have advanced English skills and are not authors of this paper. The annotators were divided into four groups that annotated the same feedback.

In particular, we randomly selected 24 essay feedback texts generated by the three task instruction types that performed best in the automatic evaluation: {\it Feedback}, {\it Feedback$\rightarrow$Scoring}, and {\it Feedback\_dCoT$\rightarrow$Scoring}. Here, we used the best-performing combination of prompt pattern and task instruction. All sampled feedback texts were written for essays from one essay set only to reduce the time the annotators need to read the essay prompt. We chose essay set 4, which covers the most common ASAP task, reading comprehension. 

To judge the feedback helpfulness, the annotators received the essay prompt, the student essay, and the generated feedback. Based on this, they were asked to assess to what extent the following statements apply on a 7-point Likert scale (score 1: ``I strongly disagree'', score 7: ``I fully agree''):

\begin{itemize}
    \setlength\itemsep{0em}
    \item[S1:] The feedback clearly points out mistakes that were made in the essay.
    \item[S2:] The feedback explains exactly why the errors are errors.
    \item[S3:] The feedback is very clear and precise so that the student can understand it.
    \item[S4:] The feedback is absolutely suitable for students from 7th to 10th grade.
    \item[S5:] Overall, the feedback is very helpful.
\end{itemize}

\paragraph{Manual Results}
Table~\ref{tab:manual-results} presents the results of the manual annotation study. For all five statements covering different helpfulness aspects, {\it Feedback} achieved the highest scores on average. Especially the clarity and precision (S3) as well as the overall helpfulness (S5) of {\it Feedback} were rated with the second-best score of 6 (``I mostly agree''). All compared task instruction types reach an average score above the neutral score of 4, indicating that all feedback is perceived as rather helpful in general. Overall, the generated essay feedback seems to have the most potential for improvement by better explaining why an error is erroneous (S2) and being more suitable for students (S4).
The inter-annotator agreement in terms of Krippendorff's $\alpha$ on average over the four groups is 0.44.

\begin{table}
    \centering
    \small
    \setlength{\tabcolsep}{4pt}
    \begin{tabular}{lrrrrr}
        \toprule
        \bf Autom. Evaluation & \multicolumn{1}{r}{\bf S1} & \multicolumn{1}{r}{\bf S2} & \multicolumn{1}{r}{\bf S3} & \multicolumn{1}{r}{\bf S4} & \multicolumn{1}{r}{\bf S5} \\
        \midrule
        Mistral & 0.29 & 0.27 & 0.45 & 0.25 & \bf 0.61 \\
        Llama-2 & --0.11 & --0.11 & --0.02 & 0.07 & --0.10 \\
        \bottomrule
    \end{tabular}
    \caption{Pearson correlation of the manual annotations per statement (S1--S5) and the automatic helpfulness scores using Mistral or Llama-2. The top value is bold.}
    \label{tab:evaluation-correlation}
\end{table}

To evaluate the reliability of our automatic helpfulness evaluation, we show the correlation between manual and automatic helpfulness scores in Table~\ref{tab:evaluation-correlation}. The highest correlation value (0.61) was measured between the manually annotated overall helpfulness (S5) and the automatic helpfulness scores predicted by Mistral. This indicates that using Mistral can be useful for automatically evaluating feedback helpfulness. The helpfulness scores generated by Llama-2 do not correlate with the manual annotation for any statement.

\section{Conclusion}
\label{sec:conclusion}

Despite the strong text generation abilities of recent LLMs in various tasks, their effectiveness in generating essay feedback that helps student writers improve their essays has remained unclear until now. Also, generating textual feedback that addresses the entire essay has previously only been tackled using one prompting strategy in a zero-shot learning setting.
With this work, we go beyond existing work by comparing different LLM prompting strategies for essay feedback generation. We propose tackling essay feedback generation and automated essay scoring (AES) jointly to study whether AES can benefit feedback generation and vice versa. Our experiments suggest that AES can be solved competitively by prompting LLMs, benefitting from tackling feedback generation first. The generated feedback is deemed helpful for students by our automatic and manual evaluation. However, the impact of scoring on the feedback helpfulness remains low ultimately.

\section{Limitations}

Aside from the still-improvable performance of the presented prompting approaches to automated essay scoring and feedback generation, we see two notable limitations of our work: 
the dependence of our feedback approaches on additional data and the pending utilization of the generated essay feedback for real-world essay writing support.

First, we point out that our feedback approaches rely on the availability of a detailed rubric, that is, guidelines including a short description for essays of each quality level, typical elements of such, and textual reasoning as to why example essays received a specific score. Such information might not always be available, which could reduce the transferability of our results to other essay datasets.

Second, while our evaluation suggests that the generated essay feedback is helpful for student writers, it remains unclear whether the student writers also perceive it as such. We encourage future work to utilize our approaches for real-world essay writing support and make it available to students. Feedback from students on such a tool would be useful to guide research on essay feedback generation.

\section*{Acknowledgments}
We would like to thank the participants of our study and the anonymous reviewers for the valuable feedback and their time. This work was partially funded by the Deutsche Forschungsgemeinschaft (DFG, German Research Foundation) within the project ArgSchool, project number 453073654.

\bibliography{bea24-aes-feedback-lit}
\bibliographystyle{acl_natbib}

\appendix
\section{Task Instructions}
\label{app:task-instructions}

We present all used task instructions in the following list. This includes all paraphrases per task instruction type.

\begin{itemize}
    \item {\bf Scoring:}    
        {\it (1)} Given this essay that was written for the given prompt, grade the essay using those ranges: \{{\it scoring\_range}\}. \\ 
        {\it (2)} Review the provided essay in response to the given prompt. Assess its quality and assign a grade according to the following criteria: \{{\it scoring\_range}\}. \\
        {\it (3)} Examine the essay written in response to the specified prompt. Utilize the following grading ranges to evaluate and score the essay: \{{\it scoring\_range}\}. \\
        {\it (4)} Analyze the submitted essay that corresponds to the given prompt. Apply these grading standards to determine its score: \{{\it scoring\_range}\}.
        
    \item {\bf Feedback:}
        {\it (1)} Analyze the given essay using the following rubric: \{{\it rubric}\}. Provide comprehensive feedback for the student that helps them to achieve better grades in the future. \\  
        {\it (2)} Please evaluate the essay in accordance with the criteria outlined in: \{{\it rubric}\}. Offer detailed and constructive feedback to assist the student in improving their writing skills for future assignments. \\
        {\it (3)} Utilize the provided rubric (\{{\it rubric}\}) to assess the essay. Your feedback should be thorough, focusing on areas of strength and suggesting improvements to help the student enhance their academic writing. \\
        {\it (4)} Conduct an assessment of the submitted essay using this specific rubric: \{{\it rubric}\}. Your feedback should be insightful and supportive, guiding the student towards achieving higher grades in their future essays. 
        
    \item {\bf Scoring$\rightarrow$Feedback:}
        {\it (1)} Grade the given essay using the following rubric: \{{\it rubric}\}. Use those score ranges: \{{\it scoring\_range}\}. Provide comprehensive feedback for the student that helps them to achieve better grades in the future. \\  
        {\it (2)} Please evaluate the essay in accordance with the criteria outlined in: \{{\it rubric}\}. Assign a grade based on these standards: \{{\it scoring\_range}\}. Offer detailed and constructive feedback to assist the student in improving their writing skills for future assignments. \\
        {\it (3)} Utilize the provided rubric (\{{\it rubric}\}) to assess the essay. Grade it according to these parameters: \{{\it scoring\_range}\}. Your feedback should be thorough, focusing on areas of strength and suggesting improvements to help the student enhance their academic writing. \\
        {\it (4)} Conduct an assessment of the submitted essay using this specific rubric: \{{\it rubric}\}. Apply the grading criteria as per these guidelines: \{{\it scoring\_range}\}. Your feedback should be insightful and supportive, guiding the student towards achieving higher grades in their future essays.
        
    \item {\bf Feedback$\rightarrow$Scoring:}
        {\it (1)} Analyse the given essay using the following rubric: \{{\it rubric}\}. Use those score ranges: \{{\it scoring\_range}\}. To do this, first provide comprehensive feedback for the student that helps them to achieve better grades in the future. Then give the final score. \\
        {\it (2)} Begin by carefully reviewing the submitted essay in light of the criteria outlined in \{{\it rubric}\}. After your thorough analysis, offer detailed and constructive feedback aimed at guiding the student towards academic improvement. Conclude your review by assigning a score to the essay, adhering to the guidelines specified in \{{\it scoring\_range}\}. \\
        {\it (3)} First, evaluate the essay against the criteria mentioned in \{{\it rubric}\}. Your evaluation should include specific, actionable suggestions for the student to enhance their writing skills and essay quality. Following your comprehensive feedback, assign a score to the essay based on the scale provided in \{{\it scoring\_range}\}. \\
        {\it (4)} Commence your assessment by applying the criteria from \{{\it rubric}\} to the essay. Focus on delivering in-depth feedback that is both informative and beneficial for the student's future academic endeavors. After providing this feedback, conclude by scoring the essay as per the range defined in \{{\it scoring\_range}\}. 

    \item {\bf Scoring$\rightarrow$Feedback\_CoT:}
        {\it (1)} Analyse the given essay using the following rubric and give helpful feedback to the student: \{{\it rubric}\}. Use those score ranges: \{{\it scoring\_range}\}. Let's think step by step. Make sure to output the score only at the end. \\
        {\it (2)} Please evaluate the provided essay according to this specific rubric: \{{\it rubric}\}. Scores should be assigned based on these criteria: \{{\it scoring\_range}\}. Proceed methodically through each step. Conclude your analysis by presenting the final score. \\
        {\it (3)} Conduct a thorough assessment of the essay using the rubric below: \{{\it rubric}\}. Adhere to the following scoring guidelines: \{{\it scoring\_range}\}. Break down your analysis into clear steps. Ensure the final score is given at the end of your evaluation. \\
        {\it (4)} Examine the student's essay in detail, utilizing the rubric provided: \{{\it rubric}\}. Apply these scoring ranges for evaluation: \{{\it scoring\_range}\}. Tackle the analysis in a step-by-step manner. The score should be presented at the conclusion of your feedback. 
    
    \item {\bf Feedback\_dCoT$\rightarrow$Scoring:}
        {\it (1)} Analyze the given essay using the following rubric and give helpful feedback to the student: \{{\it rubric}\}. Use those score ranges: \{{\it scoring\_range}\}. Let's think step by step. First, analyze the quality of the essay in terms of the given rubric. Then, give feedback to the student that explains their mistakes and errors and additionally gives them tips to avoid them in the future. As a final step, output the score at the end. \\ 
        {\it (2)} Begin by evaluating the essay based on the criteria outlined in the rubric: \{{\it rubric}\}. Consider the scoring guidelines provided: \{{\it scoring\_range}\}. First, conduct a thorough analysis of the essay according to the rubric standards. Next, provide constructive feedback to the student, highlighting areas for improvement and suggesting strategies to enhance their writing skills. Conclude with a summary of the essay's strengths and weaknesses. Finally, present the essay's score at the end of your analysis. \\
        {\it (3)} Follow these steps to assess the student's essay: First, reference the provided rubric: \{{\it rubric}\}, and apply it to evaluate the essay. Use the scoring ranges given: \{{\it scoring\_range}\} for accurate assessment. Provide detailed feedback to the student, pinpointing specific areas of the essay that align or deviate from the rubric, along with advice for future improvement. Your feedback should be clear, constructive, and actionable. After your comprehensive review, conclude by outputting the final score, ensuring this is done only at the very end. \\
        {\it (4)} To evaluate the student's essay, proceed as follows: Start with the provided rubric: \{{\it rubric}\}, to assess the essay's attributes. Adhere to the scoring guidelines: \{{\it scoring\_range}\} for consistency. Your analysis should first focus on how well the essay meets the criteria in the rubric. Then, craft feedback for the student that is both informative and helpful, addressing any shortcomings and providing practical advice for future essays. The feedback should be encouraging yet honest. Conclude your evaluation by scoring the essay, presented at the conclusion of your feedback. 
        
    \item {\bf Scoring$\rightarrow$Explanation:}
        {\it (1)} Grade the given essay using the following rubric: \{{\it rubric}\}. Use those score ranges: \{{\it scoring\_range}\}. Provide an explanation for your score as well. \\
        {\it (2)} Please assess the submitted essay according to the criteria outlined in this rubric: \{{\it rubric}\}. Scores should be allocated based on these guidelines: \{{\it scoring\_range}\}. Additionally, include a detailed rationale for the score you assign. \\
        {\it (3)} Evaluate the provided essay by referring to the standards specified here: \{{\it rubric}\}. Utilize the following scoring range for your evaluation: \{{\it scoring\_range}\}. Also, furnish a comprehensive justification for the grade you determine. \\
        {\it (4)} Rate the essay in front of you using these evaluation criteria: \{{\it rubric}\}. Your scoring should align with these parameters: \{{\it scoring\_range}\}. Please also give a thorough explanation to support the score you decide upon. 
        
    \item {\bf Explanation$\rightarrow$Scoring:}
        {\it (1)} Analyse the given essay using the following rubric: \{{\it rubric}\}. To do this, first explain using the scoring rubric why you chose the score. After you analysed the essay, give a final grade. \\
        {\it (2)} Utilize the provided scoring rubric (\{{\it rubric}\}) to evaluate the essay. Begin by detailing the reasons for your assigned score based on the rubric's criteria. Conclude by summarizing your analysis with a final grade. \\
        {\it (3)} Apply the scoring guidelines from \{{\it rubric}\} to assess the essay. Start by discussing how the essay meets or falls short of each criterion in the rubric. Finalize your assessment with a grade. \\
        {\it (4)} Use the scoring rubric (\{{\it rubric}\}) as a basis to critically analyze the essay. Explain how the essay aligns with each aspect of the rubric, justifying your evaluation. Conclude with a definitive grade. 
\end{itemize}

\section{Automatic Helpfulness Assessment}
\label{app:automatic-helpfulness}

We used the following prompt for our automatic helpfulness evaluation:

``You are given an essay and feedback from a teacher for this essay. Your task is to evaluate the helpfulness of the feedback.

\# Task:
Evaluate the helpfulness of the feedback. Helpful feedback should explain what the errors are, why they are errors, and how to fix them. Give a score between 1 and 10, where 1 means the feedback is not helpful at all, and 10 means the feedback is very helpful.

Provide the output in the following output:
\{format\_instructions\}''

\section{Generated Essay Feedback}
\label{app:generated-essay-feedback}

\begin{table*}[t]
    \centering
    \small
    \setlength{\tabcolsep}{2pt}
    \begin{tabular}{lp{12.25cm}}
        \toprule
        \bf Task Instruction Type & \bf Generated Essay Feedback \\
        \midrule
        Feedback & Based on the provided essay prompt and the given text, it appears that the student has not fully understood the story "Winter Hibiscus" by Minfong Ho. The student's response, "Saeng would return to her homeland," is not an accurate reflection of the story's conclusion.

The author concludes the story with Saeng's determination to retake the driver's test in the spring when the hibiscus is budding and the geese have returned. This decision comes after Saeng's emotional experience in the greenhouse, where she is surrounded by familiar plants and memories, and is struck by a wave of loss for her past life in Vietnam.

The author's use of the imagery of the geese returning and the hibiscus budding signifies the cyclical nature of life and the passing of seasons. Saeng's decision to retake the driver's test in the spring shows her resilience and determination to move forward in her new life in the United States while holding on to her past.

To improve future essays, the student could focus on closely reading and understanding the text, taking notes, and identifying key themes and symbols. It may also be helpful for the student to practice summarizing and paraphrasing the text in their own words to demonstrate a deeper understanding. \\
        \midrule
        Feedback$\rightarrow$Scoring & The essay response is very brief and does not fully address the demands of the question. The statement 'Saeng would return to her homeland' is not explicitly stated in the text, and the essay does not provide any evidence or explanation to support this assertion. To improve, consider reading the text more closely and taking notes to help you understand the main ideas and themes. Use these notes to develop a more detailed and well-supported response that addresses the demands of the question. \\
        \midrule
        Feedback\_dCoT$\rightarrow$Scoring & Your essay does not fully address the demands of the question. In order to earn a higher score, you should explain why Saeng would return to her homeland based on the details and events in the story. Consider how the hibiscus plant and the memories it evokes for Saeng might be connected to her desire to go back to her homeland. Additionally, you could discuss how the experiences and emotions Saeng encounters in the greenhouse contribute to her decision. Remember to use specific details and examples from the text to support your ideas.\\
        \bottomrule
    \end{tabular}
    \caption{Generated essay feedback using the three task instruction types compared in the manual annotation study. The feedback was generated for the same student essay from essay set 4 of the ASAP dataset \cite{hamner-etal-2012-hewlett}.}
    \label{tab:examples}
\end{table*}

Table~\ref{tab:examples} compares the generated essay feedback for the same essay using three different task instruction types, compared in the manual evaluation.

\end{document}